\documentclass[10pt,letterpaper]{article}
\usepackage{opex3}
\usepackage{color}
\usepackage{subfig}

\usepackage{epstopdf}
\usepackage{amsmath}
\usepackage{amssymb}
\usepackage{mathtools}
\usepackage{cite}

\DeclarePairedDelimiter\floor{\lfloor}{\rfloor}

\DeclareMathOperator*{\argmin}{arg\,min}

\newcommand{\etal}{{\emph{et~al. }}}
\newcommand{\Section}[1]{\vspace{-4pt}\section{#1}\vspace{-4pt}}

\newcommand{\Subsection}[1]{\vspace{-3pt}\subsection{#1}\vspace{-3pt}}
\newenvironment{tight_itemize}{\begin{itemize} \itemsep
		-3pt}{\end{itemize}}

\begin{document}

\title{Spatial Phase-Sweep: Increasing temporal resolution of transient imaging using a light source array}

\author{Ryuichi Tadano$^{1,*}$, Adithya Kumar Pediredla$^2$, Kaushik Mitra$^3$ and Ashok Veeraraghavan$^2$}

\address{$^1$Sony Corporation, 7-1 Konan 1-chome, Minato-ku, Tokyo 108-0075, Japan\\
$^2$Rice University, 6100 Main Street, Houston, TX 77005, USA\\
$^3$Indian Institute of Technology Madras, Chennai, Tamil Nadu 600036, India}

\email{$^*$Ryuichi.Tadano@jp.sony.com} 



\begin{abstract*}
Transient imaging or light-in-flight techniques capture the propagation of an ultra-short pulse of light through a scene, which in effect captures the optical impulse response of the scene. 
Recently, it has been shown that we can capture transient images using commercially available Time-of-Flight (ToF) systems such as Photonic Mixer Devices (PMD). 
In this paper, we propose `spatial phase-sweep', a technique that exploits the speed of light to increase the temporal resolution beyond the 100 picosecond limit imposed by current electronics. 
Spatial phase-sweep uses a linear array of light sources with spatial separation of about 3 mm between them, thereby resulting in a time shift of about 10 picoseconds, which translates into 100 Gfps of transient imaging in theory. 
We demonstrate a prototype and transient imaging results using spatial phase-sweep.
\end{abstract*}

\vspace{10pt}


\bibliographystyle{osajnl}

\Section{Introduction}
Transient imaging or light-in-flight refers to capturing the temporal response of a scene to an ultra-short pulse of light. 
Current techniques to capture transient images are either based on streak cameras or on photonic mixer devices. 
Streak cameras when used along with femtosecond laser pulse based illumination can provide extremely fine temporal resolution, of the order of 1 picosecond, but such systems \cite{Velten2011,Velten2013,Heshmat2014,Velten2012} are prohibitively expensive and cost upwards of several hundred thousand dollars. 
More recently, Heide \etal \cite{Heide2013}, Kadambi \etal \cite{Kadambi2013}, and O'Toole \etal \cite{OToole2014} have shown that commercially available photonic mixer devices that cost a few hundred dollars can be used to acquire transient images.
Unfortunately, the temporal resolution of these techniques is limited by the accuracy of the phase locked loop (PLL) circuit in the on-board electronics of these devices. 
In commercially available systems such as camboard nano \cite{camboardnano}, the on-board electronics and the PLL limits the minimum achievable phase shift to the order of about 100 picoseconds ($\sim$128 picoseconds on camboard nano). 
As a consequence, transient images obtained using photonic mixer devices have a much lower temporal resolution (100 picoseconds) compared to systems based on streak cameras and femtosecond laser pulses (1 picosecond). 

\begin{figure}[t]
	\begin{center}
		\subfloat[Conventional system \cite{Kadambi2013}]{
			\includegraphics[width=2.2in]{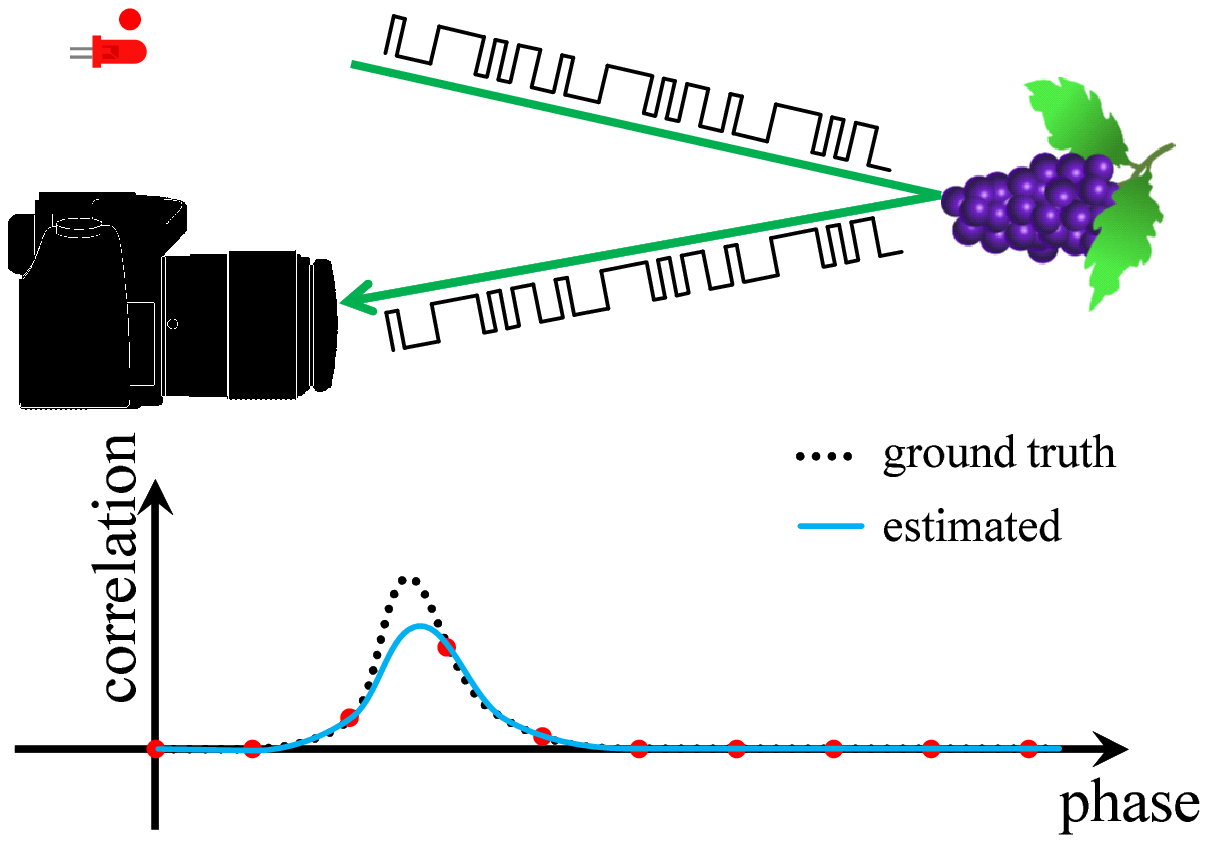}
		}
		\hspace{0.3in}
		\subfloat[Spatial Phase-Sweep (proposed)]{
			\includegraphics[width=2.2in]{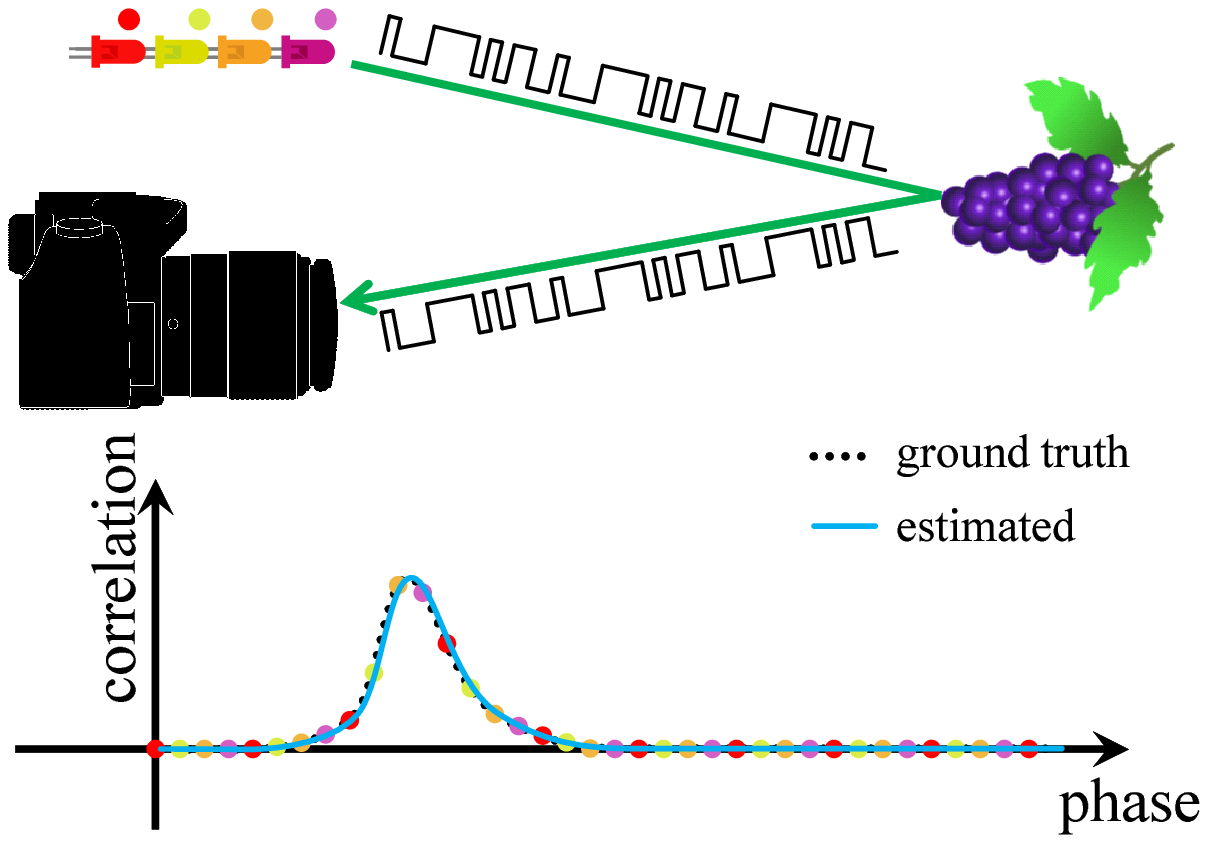}
		}
	\end{center}
	\caption{(view in color)
		{\bf Schematic chart of Spatial Phase-Sweep:}
		To capture the impulse response of the scene, we employ a pseudo random code, whose auto correlation is triangle shaped.
		(a) The sampling rate of auto correlation is limited by the minimum phase shift amount of PLL circuit in conventional systems \cite{Kadambi2013}. 
		(b) In our system, using the same electronics, a light source array is utilized to introduce more samples in phase by means of spatial domain sweep.
	}
	\label{fig:concept}
\end{figure}

Our goal in this paper is to improve the temporal resolution of transient images obtained using photonic mixer devices (PMD) over and above the limit imposed by the sensor electronics. 
We exploit the incredible speed of light ($3\times10^8$ m/sec) to our advantage and propose a technique called `spatial phase-sweep' (SPS) to improve the temporal resolution of transient images obtained using PMDs. 
The idea behind spatial phase-sweep is very simple. 
We use an array of light sources with the different sources in the array being slightly offset along the optical axis.
That creates small but precisely controllable differences in time of travel between the light pulses emitted by the different sources (Fig.~\ref{fig:concept}).
Since the light source positions in an array can be precisely controlled, the corresponding path length differences created result in a slight temporal offset --- the temporal offset $\Delta t$, is given by $\Delta t = \frac{\Delta d}{c}$, where $\Delta d$ is the spatial shift between adjacent light sources in the array, and $c$ is the speed of light.
In our prototype, $\Delta d$ is 3 mm, resulting in a temporal resolution $\Delta t$ of about 10 picoseconds, an order of magnitude better than the limit imposed by the on-board electronics of the PMD device in the prototype.

The main alternative technique to improving the temporal resolution of transient images obtained using photonic mixer devices, is to increase the base frequency of the voltage controlled oscillator (VCO) used in the phase locked loop circuit in the on-board electronics. 
Boosting the base frequency of the VCO may theoretically provide up to 10x improvement in the temporal resolution, but such a technique would come with significant increase in the cost of the resulting sensor. 
Our proposed technique results in minor incremental cost over existing solutions, since we only need to create a linear array of laser diodes, which are inexpensive and easy to obtain. 
In addition, the key innovation of spatial phase-sweep is independent of the temporal resolution limit imposed by on-board electronics. 
This means that even if sensor electronics are improved significantly, the spatial phase-sweep technique may be used to further improve temporal resolution over that limit. 
The fundamental limit on temporal resolution achieved using spatial phase-sweep is dependent mainly upon the accuracy with which one can control the positioning of laser diodes in the array. 
Since the positioning of laser diodes can be controlled to sub-millimeter precision, spatial phase-sweep will continue to provide improvement in temporal resolution even if the on-board electronics improve by an order of magnitude, due to increased base frequency of the VCO. 

The main technical contributions of our paper are as follows:
\begin{tight_itemize}
	\item{We propose spatial phase-sweep, a technique to improve the temporal resolution of transient images captured using photonic mixer devices.}
	\item{We develop algorithms for self-calibration and transient imaging recovery from the data captured using spatial phase-sweep ToF camera.}
	\item{We build a proof of concept prototype and demonstrate a 10x improvement in temporal resolution}
\end{tight_itemize}

Some of the limitations of the proposed technique are:
\begin{tight_itemize}
	\item{The data acquisition time increases linearly with the increase in temporal resolution.}
	\item{The physical size of the light sources will limit the size of the light source array and hence, the increase in the resolution of spatial phase-sweep beyond a limit will be difficult to implement.}
	\item{Our system requires repetitive measurements, which means that it cannot capture one-time phenomena such as plasma dynamics.}
\end{tight_itemize}

\Section{Prior work}
\label{sec:prior_work}
Transient imaging finds applications in visualizing the interaction of light with an optically complex scene that can involve multiple reflections, scattering media, or subsurface scattering.
In this section, we review various approaches to transient imaging and proceed to explain about our  approach. 

\noindent \textbf{Holography based:}
Abramson captured the first light-in-flight images by shining a flat surface and a hologram with a short laser pulse \cite{Abramson1978,Abramson1983}. 
The beam from the flat surface is used as reference beam and the light coming from hologram interferes with the reference beam to produce an image that corresponds to a short distance traveled by the light wave. 
By moving the reference surface and stacking the images, they created light-in-flight images. 
Nilsson \cite{Nilsson1998} repeated the same experiment with the help of CCD array to create digital light-in-flight video.

\noindent \textbf{OCT based:}
Gkioulekas \etal \cite{Gkioulekas2015} proposed micron-scale transient imaging using optical coherence tomography (OCT).
The idea of incorporating OCT technique is close to ours, however the scale of the subject they support is quite small, 2 cm H $\times$ 2 cm W $\times$ 1 cm D.

\noindent \textbf{Streak camera based:}
Velten \etal \cite{Velten2011,Velten2012,Velten2013} proposed the use of a streak camera and a femtosecond laser to capture transient images. 
The laser illuminates one horizontal scan line at a time and scans the entire scene. 
For every scan, photons illuminate the scene, scatter and some of the scattered photons eventually reach the streak camera. 
The streak camera converts these photons into electrons using a photo cathode.
These electrons are then deflected vertically by a voltage that varies with time. 
Hence, the intensity of the pixels in the vertical axis of the image correspond to the photons coming from various depths. 
Scanning the entire scene, Velten \etal  produced high resolution transient image ($\sim$1 picosecond).



The need for scanning in this approach makes it difficult to handle non-repetitive time-evolving events, such as laser ablation, optical rogue waves, sonoluminescence, and nuclear explosion.
To solve this problem, Gao \etal \cite{Gao2014} employed digital micro-mirror device (DMD) and compressed sensing techniques along with streak camera. 
Their system achieved a temporal resolution of 10 picosecond. 
Heshmat \etal \cite{Heshmat2014} utilized a tilted lenslet array to realize a single shot transient imaging at a temporal resolution of 2 picosecond. 


\noindent \textbf{PMD based:}
Though streak camera based methods provide very high temporal resolution, they are prohibitively expensive: a system based on femtosecond laser and a streak camera costs upward of several hundred thousand dollars.
To realize an inexpensive transient imaging, photonic mixer device (PMD) based methods have been proposed by Heide \etal \cite{Heide2013}.

PMDs are the basic building blocks of most commercial time of flight cameras.
Besides, several applications using this device have been proposed in the past few years \cite{Kadambi2013,Heide2014,Tadano}.
In such systems, a laser diode or a light emitting diode (LED) is temporally modulated to create a coded illumination signal. 
The light scattered off the subject is then correlated with a programmable sensor modulation pattern on a PMD sensor to obtain an array of correlational measurements.
Heide \etal \cite{Heide2013} performed a series of measurements with varying phase delays between the illumination and the sensor modulation patterns (while keeping both to be sinusoidal), and demonstrated a deconvolution technique that is capable of recovering the transient images from the captured correlational measurements.
Kadambi \etal \cite{Kadambi2013} demonstrated a similar technique for recovering transient images, but using M-sequence, instead of sinusoidal modulation.

O'Toole \etal \cite{OToole2014} used an encoded projector to modulate the light both spatially and temporally.
The 3-D illumination signal is transformed by interacting with the scene and is captured by the PMD sensor. 
The spatial and the temporal components of the received signal have complementary information about the scene and are used to more robustly capture sharp light-in-flight images. 
The temporal resolution of their transient image is 100 picoseconds, which translates to 10 Gfps.

All these PMD based techniques for capturing transient images using PMD sensors are limited in their temporal resolution, primarily due to phase locked loop (PLL) in FPGA or electrical circuits.
PLLs in these commercially available electrical circuits are limited to about 100 picosecond time delays, which results in a 100 picosecond temporal resolution on the captured transient images. 
In this paper, we will overcome this limit through spatial phase-sweep, while restricting the cost of the device to be low.



\Section{Background}
\label{sec:transient_image_using_tof}
\begin{figure}[t]
	\begin{center}			
		\subfloat[ToF camera system]{
			\includegraphics[width=2.9in]{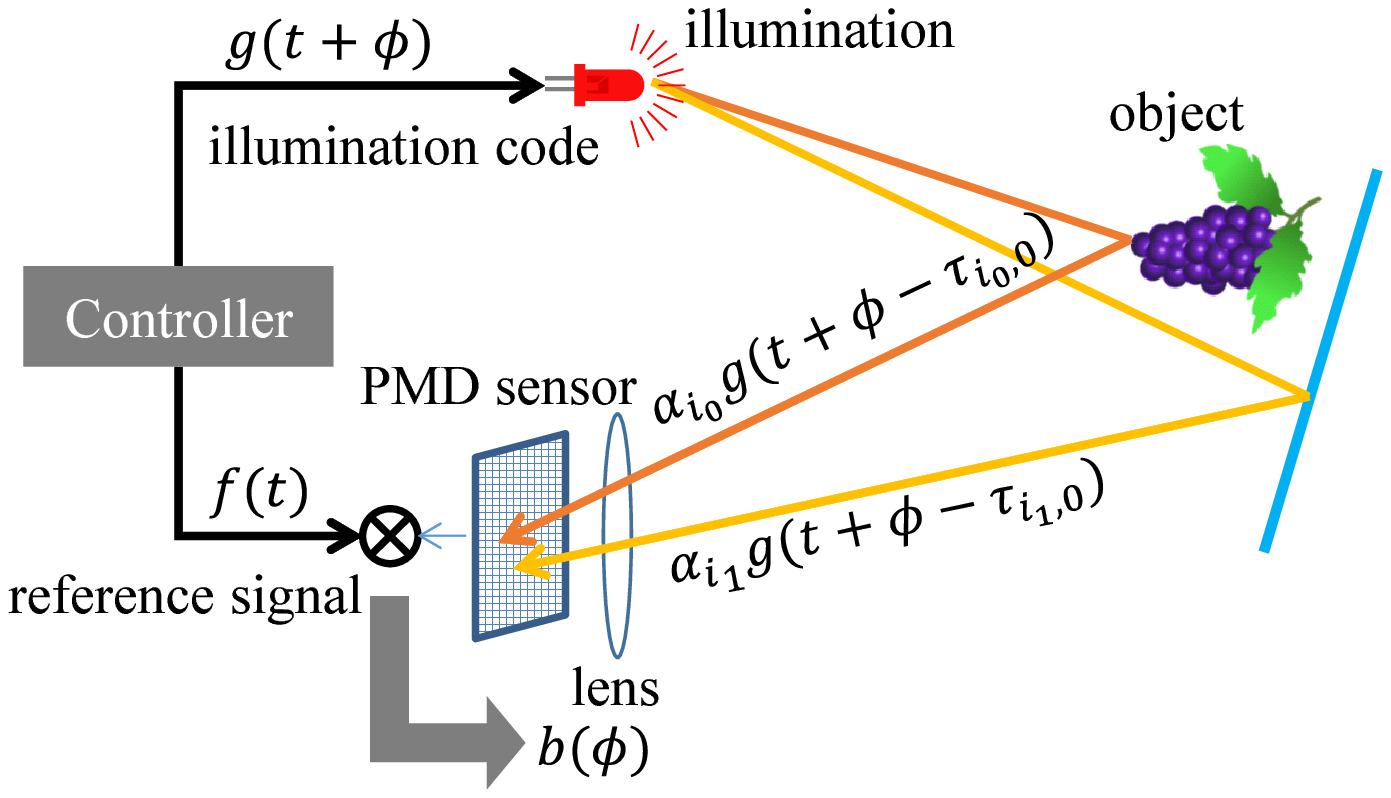}
			\label{fig:tof_system}
		}
		\subfloat[Auto correlation of m-sequence (31 bits).]{
			\includegraphics[width=2.3in]{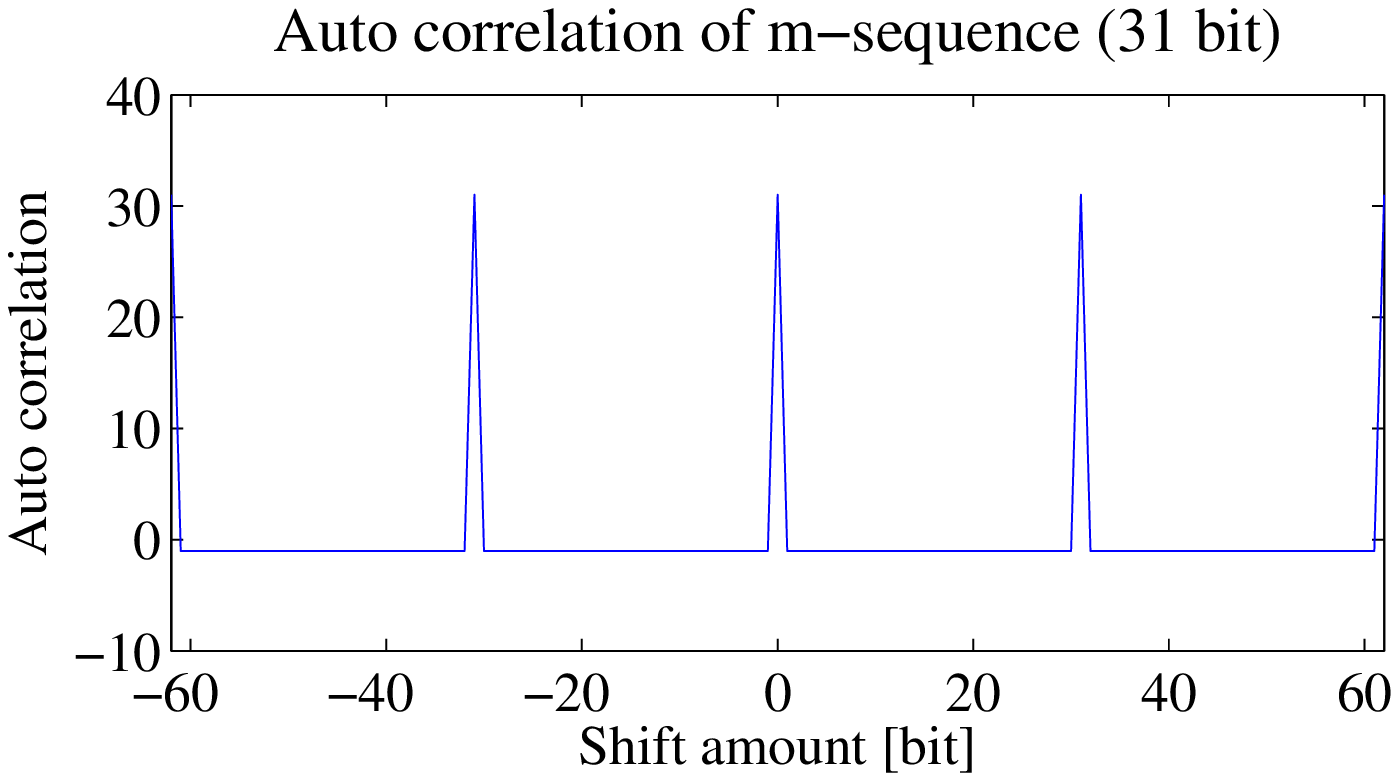}
			\label{fig:acorr_mseq}
		}
	\end{center}
	\caption{
		(a) {\bf ToF camera system:}
		The system controller sends two binary signals: $f(t)$ to the PMD sensor and $g(t+\phi)$ to the illumination source. 
		For each pixel, the PMD sensor measures correlation between reference signal and incident light on pixel (reflected version of $g(t+\phi)$) as shown in Eq.~\ref{eq:pmd_observation}.
		(b) {\bf Auto correlation of m-sequence:}
		We use m-sequence for both $f(t)$ and $g(t)$.
		The interval between each peak in the auto correlation of m-sequence can be controlled by the code length. 
		Using a sufficiently long code, we can focus on only one peak and treat it as delta function. 
	}
	\label{fig:system_and_code_design}
\end{figure}
In this section, we will explain the principles of PMD sensor based ToF camera and proceed to explaining on how ToF camera can be used to obtain transient images. 
\Subsection{Time-of-Flight principles}
ToF camera \cite{Moller2005,Lange2000,Lange2001,Schwarte1997,Conroy2009} consists of a PMD sensor and laser diodes that emit coded illumination $g(t)$. 
This illumination signal interacts with the scene and reaches a sensor pixel. 
Due to the available technology, the sensor cannot directly measure the signal received, but can only measure the correlation between the received signal and a binary coded signal $f(t)$ inside the sensor circuit. 
The entire process for each pixel can be mathematically represented as
\begin{equation}
b(\phi)=\int_{0}^{T}\alpha(\tau)\cdot\int_{0}^{\infty}g(t+\phi-\tau)f(t)\,dt\,d\tau,\quad{\rm with}\quad\alpha(\tau)=\int_{p}\alpha_{p}\delta(|p|=\tau),
\label{eq:pmd_observation}
\end{equation}
where $\tau$ is temporal delay of illumination due to the finite speed of light that travels from the light source to the sensor pixel via scene,
$\alpha(\tau)$ is the scene response (integration of all contributions from different light paths $p$ that correspond to the same delay $\tau$),
$T$ is exposure time,
and $\phi$ is delay for illumination signal controlled by the system.

In an ordinary ToF camera that is designed to capture depth information, it is assumed that the scene has just single path. Hence, Eq.~\ref{eq:pmd_observation} becomes 
\begin{equation}
b(\phi)=\alpha\cdot\int_{0}^{\infty}g(t+\phi-\tau_0)f(t)\,dt.
\label{eq:pmd_observation_single} 
\end{equation}
Where, $\tau_0$ is the time delay.
Further, sinusoidal waves with same frequency are utilized for both $f(t)$ and $g(t)$. 
Three or four measurements with different amounts of phase shift are required to generate depth information. 
When multiple cameras are in operation, custom codes such as pseudo random sequence are utilized to overcome interference problems \cite{Buttgen2008, Whyte2010}.

\Subsection{Transient imaging using Time-of-Fight camera}
$\alpha(\tau)$ in Eq.~\ref{eq:pmd_observation} is the impulse response of the world or transient response that we are interested to solve, not just for the single path case, but for a generic case. 
The most common approach to solve for $\alpha(\tau)$ is by de-convolving $b(\phi)$ with cross-correlation function between $f(t)$ and $g(t)$ \cite{Heide2013, Kadambi2013} .
In \cite{Heide2013}, various combinations of frequency/phase-shifted sinusoidal functions are used for $f(t)$ and $g(t)$ to build a correlation matrix.
To solve the inverse problem described by the matrix, they incorporated regularization functions restricting the transient response to be smooth in temporal and spatial domain. 
In \cite{Kadambi2013}, $f(t)$ and $g(t)$ are designed to be m-sequences so that inverting the cross-correlation function becomes easy. 

The common problem with both the approaches is that the measurements $b(\phi)$ cannot be sampled at arbitrary sampling rate.
With the existing electronics, $b(\phi)$ can only be sampled at 10 Gfps. 
Light events such as subsurface scattering or inter-reflections happen at much faster rate and are missed in these approaches. 
Hence, $b(\phi)$ is grossly under sampled. 
We propose to increase the sampling rate of measurement vector $b(\phi)$ by a factor of 10, thereby capture these fast occurring transient events more accurately, at 100 Gfps.
Note that $b(\phi)$ cannot be described in a parametric way as it includes an unknown scene response.
Hence, the only way to improve temporal resolution of transient image is to do finer sampling.

\begin{figure}[t]
	\centering\includegraphics[width=4.8in]{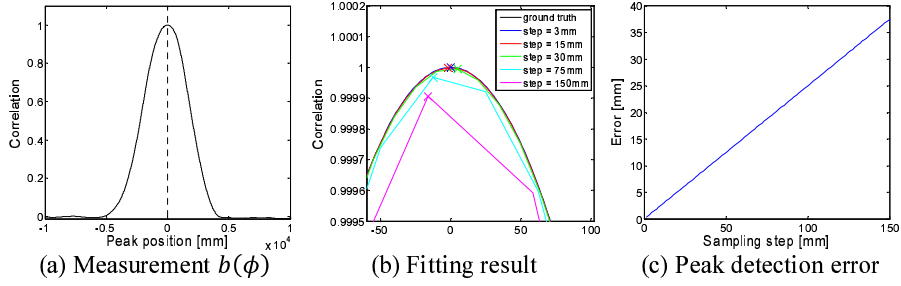}
	\caption{
		{
		\textbf{Sampling step and peak detection error:}
		(a) Measurement $b(\phi)$. Due to subsurface scattering/indirect reflections, actual cross correlation does not look like a triangular form.
		(b) OMP based kernel fitting results.
		Estimated peak positions are marked as X. 
		Smaller sampling step better fits the estimated curve to ground truth.
		(c) Relationship between sampling step and peak estimation error. 
		Error increases as the sampling step becomes larger. 
		}
	}
	\label{fig:sampling_step_and_fitting_error}
\end{figure}
To show that, we perform a simple simulated experiment based on an actual measurement $b(\phi)$.
Here, we followed the transient imaging method of \cite{Kadambi2013}, which uses m-sequence as $f(t)$ and $g(t)$.
Using a measurement $b(\phi)$ as ground truth, we performed OMP based kernel fitting \cite{Kadambi2013}. 
In the OMP based kernel fitting, sub-sampled data of ground truth is used as kernel basis.
For simplicity, we assume the scene response is 1-sparse in terms of sub-sampled kernel basis.
We change the interval of sampling points to investigate how sampling step affects the fitting results.
As shown in Fig.~\ref{fig:sampling_step_and_fitting_error}~(a), the actual measurement is not triangular shaped as expected from Fig.~\ref{fig:system_and_code_design}~\subref{fig:acorr_mseq}.
This unknown shape is difficult to describe in a parametric way.
OMP based fitting will provide us better estimation because it is based on actual measured kernel.
Even for a simple task such as peak estimation, we can see that increasing sampling rate gives us a finer estimation results (Fig.~\ref{fig:sampling_step_and_fitting_error}~(b), (c)).
Hence, doing finer sampling will increase the information we can obtain, especially for complicated tasks.

\begin{figure}[t]
	\centering\includegraphics[width=4.5in]{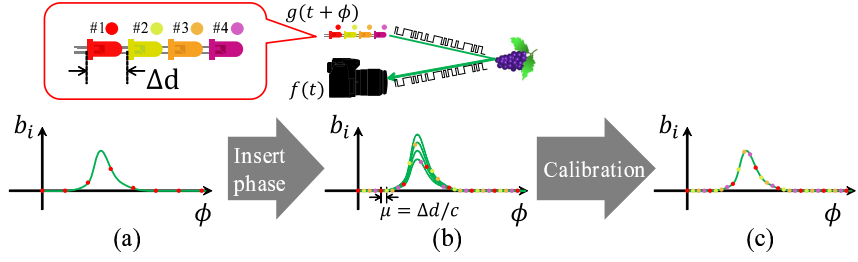}
	\caption{
		\textbf{Spatial phase-sweep:} 
		By placing multiple light sources in an array with slightly different distances from the subject, we can sample the cross correlation function more precisely than the conventional PLL based sampling. 
		(a) Graph showing measurements only from light source \#1. 
		(b) The observations for each light source are taken independently. 
		Note that amplitude of those data differ due to the variation of distance or the angle at which the light reaches. 
		(c) We combine the data from different light sources by performing equalization.
		}
	\label{fig:phase_insertion}
\end{figure}
\Section{Increasing temporal resolution of transient imaging}
\label{sec:increasing_temporal_resolution}
As described in Sec.~\ref{sec:prior_work}, temporal resolution of conventional transient imaging techniques using PMD sensor are theoretically limited to around 100 picoseconds \cite{Kadambi2013}. 
This is determined by the precision of phase shift control ($\phi$) of the PLL circuit. 
In this section, we first formulate how the precision of controlling $\phi$ affects the information we can acquire. We then introduce a simple technique to boost the temporal resolution without increasing the phase shift precision control of PLL. 
The concept of our idea is illustrated in Fig.~\ref{fig:phase_insertion}.

\Subsection{Spatial phase-sweep}
Suppose $h(x)$ is the cross correlation function between $f(t)$ and $g(t)$, Eq.~\ref{eq:pmd_observation} can be written as
\begin{equation}
b(\phi)=\int_{0}^{T}\alpha(\tau)h(\phi-\tau)d\tau,\quad{\rm with}\quad h(x)=\int f(t)g(t+x)dt.
\label{eq:pmd_observation_simple}
\end{equation}
Hence, solving for transient response is a deconvolution problem and is more intuitive in frequency domain. Computing the discrete-time Fourier transform on both sides of Eq.~\ref{eq:pmd_observation_simple} and rearranging the terms, we have
\begin{align}
	{{\mathcal{A}}}{ (2\pi f \Delta \phi)} = \frac{\mathcal{B}(2\pi f \Delta \phi)}{\mathcal{H}(2\pi f \Delta \phi)}
	\label{eq:dtft}
\end{align}


where $\Delta\phi$ is sampling interval (or phase shift amount), and $\mathcal{A}$, $\mathcal{B}$, and $\mathcal{H}$ are discrete-time Fourier transforms of $\alpha$, $b$ and $h$.
Note that sampling performed here is not in temporal domain but in phase domain.
Clearly, Eq.~\ref{eq:dtft} is periodic with period $\frac{1}{\Delta\phi}$. 
Hence, smaller $\Delta\phi$ is better as we can capture more frequency information without aliasing.
For the commercially available PLL circuits, the phase shift control ($\Delta \phi$) is around 100~ps.
However, light events such as sub-surface scattering happen at much higher frequency, depending on the properties of the material. 
Hence, it is crucial to have smaller sampling interval $\Delta \phi$ to acquire more information about transient image.

The currently available oscillator's frequency of 100 ps corresponds to a distance of 3 cm traveled by light. 
With the current state of the art design, the 3 cm precision control determines the theoretical limitation of the frequency of the transient image. 
To break this limit, we insert extra phase delay that is independent from the innovations in PLL's design, by arranging an array of light sources uniformly and perpendicular to the image plane. 
We call this idea as `Spatial phase-sweep' as the spatial arrangement of light sources sweeps the phase of illumination signal. 
After incorporating the extra freedom of the position of the light source in Eq.~\ref{eq:pmd_observation}, the measurements are re-formulated as:
\begin{eqnarray}
b(\phi + \mu_n)=&\displaystyle \int_{0}^{T}\alpha(\tau)\cdot\int_{0}^{\infty} g(t+\phi+\mu_{n}-\tau)f(t)\,dt\,d\tau, 
\label{eq:pmd_observation_multiple}
\end{eqnarray}
where $\mu_{n}$ is phase delay inserted by $n^{\scriptsize \mbox{th}}$ light source. Though the light sources can be arbitrarily placed, we place them uniformly. Hence, $\mu_{n}$ is given by $ \mu_{n}=n\cdot\frac{\Delta d}{c}$, where $\Delta d$ is the distance between two consecutive light sources. 
In summary, we can now sample the measurements at the rate of $ \frac{\Delta d}{c}$ = 10 ps,
allowing us to acquire information that is previously not possible.




\Subsection{Calibration for phase insertion}
As we change the active light source the amplitude of incident light at each pixel and the distance between object and light source changes. 
To overcome this inconsistency, we introduce an equalization process between the data taken with different illumination sources. 
Let us call measurements set for multiple light source positions as $\{b_{n}(\phi)\}$, where $n$ denotes the index of active light source. 
For each measurement $n$, we calculate equalization coefficient $w_{n}$ by minimizing the following cost function via least squares method:
\begin{align}
w_{n}=&\displaystyle \argmin_{w}\,\,\sum_{\phi}\left|\left|\hat{b}_{n}(\phi)-w\cdot b_{n}(\phi) \right|\right|_{2}^{2} \nonumber \\
=&\displaystyle \frac{\sum_{\phi}(\hat{b}_{n}(\phi)\cdot b_{n}(\phi))}{\sum_{\phi}(b_{n}^{2}(\phi))}
\end{align}

where $b_{n}(\phi)$ is measurement corresponding to $n^{\scriptsize \mbox{th}}$ light source, and $\hat{b}_{n}(\phi)$ is estimation of equalized $b_{n}(\phi)$ obtained by linearly interpolating data set $\{b_{0}(\phi)\}$. The cost function is intended to decrease the squared error between $\hat{b}_{n}(\phi)$ and equalized observation $w\cdot b_{n}(\phi)$. Fig.~\ref{fig:phase_insertion} illustrates the basic idea of Spatial Phase-Sweep.



\begin{figure}[h]
	\centering\includegraphics[width=\textwidth]{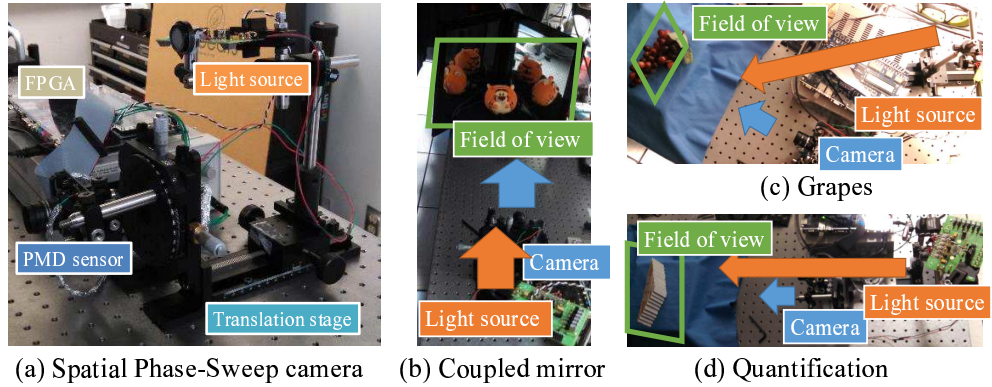}
%
	\caption{
		(color in electronic version) \textbf{System implementation and scene setup:}
		(a) Our implementation is comprised of Altera FPGA development kit DE2-115, infrared laser diode, PMD 19k-S3, and a translation stage.
		We show the effect of our spatial phase-sweep technique on three scenes: 
		an object placed between a coupled mirror (b),
		grapes which includes small spherical surfaces (c), and
		quantification scene which includes stacked 10 sheets of 3 mm thickness (d).
		In each picture of setups, the directions faced camera and light source are described with orange and blue arrows. 
		Also, green frames indicate the actual field of view which the camera sees.
	}
	\label{fig:scene_setup}
\end{figure}

\Section{Experimental Setup}
\label{sec:experiments}
Our system consists of a PMD sensor, laser diode, Altera’s FPGA development kit DE2-11 and a translation stage. 
To simulate light source array, we use a translation stage to control the light source position linearly towards the subject. 
Fig.~\ref{fig:scene_setup}~(a) shows our setup. FPGA controls various functions of PMD sensor including the reference code $f(t)$. 
Captured image is read out via FPGA and saved in an external storage. 
FPGA also controls laser diode driving board by sending illumination code $g(t)$. 
This ensures that the frequency and phase of the light source and sensor are synchronized.
Most of the hardware and software design of our system is based on the work by Kadambi \etal \cite{Kadambi2013}.

\textbf{Illumination:}
The infrared laser diode in our set up is used for illumination. 
The diodes are driven by iC-HG from iC-Haus. We can choose arbitrary binary sequence as illumination code $f(t)$. 
In our system, we used a 31 bit m-sequence. 
The modulation frequency was 50 MHz. 
As mentioned above, we changed a single light source position for each measurement by a translation stage to simulate a light source array. 
Using such a mechanical component, we can control $\Delta d$ of Eq.~\ref{eq:pmd_observation} in the orders of 0.1 mm.
%
%

\textbf{Code control:}
\label{sec:code_control}
The PLL circuit included in the FPGA allows us to shift the phase of the output signal depending on the VCO frequency. 
In our configuration, we can control $\phi$ by about 96 ps. 
This is more precise compared to the code modulation frequency. 
This phase shift amount corresponds to light travel distance of about 2.8 cm, which implies that the frame rate of the transient image we can get is around 10 Gfps.

\textbf{Translation stage:}
In our experiment, we utilized linear translation stage to control the position of the light source in the orders of 0.1 mm. 
However, for the approximations given in \ref{subsection:Systematicerror} to be valid, we used a step size of 2.8 mm. 
Hence, the translation stage helped us in inserting 9 extra measurement to increase the temporal resolution 10 times, to 100 Gfps.

\begin{figure}[t]
	\begin{center}
		\subfloat[Schematic chart for qualification]{
			\includegraphics[width=1.8in]{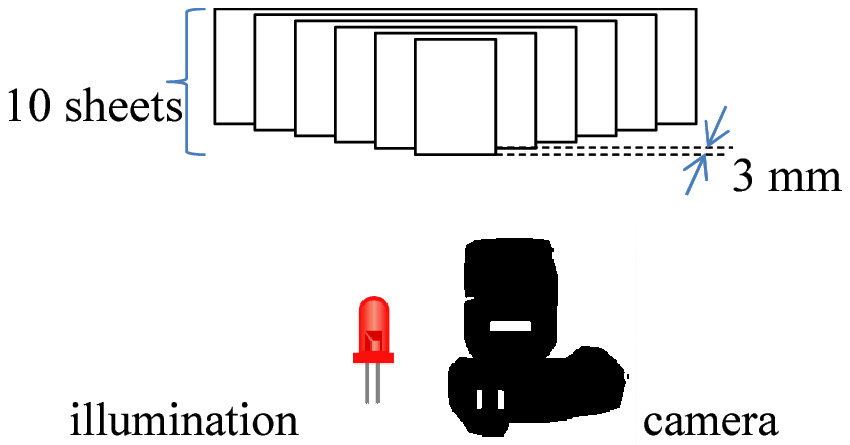}
			\label{fig:qualification_evaluation_setup}
		}
		\subfloat[Temporal resolution \cite{Kadambi2013}]{
			\includegraphics[width=1.5in]{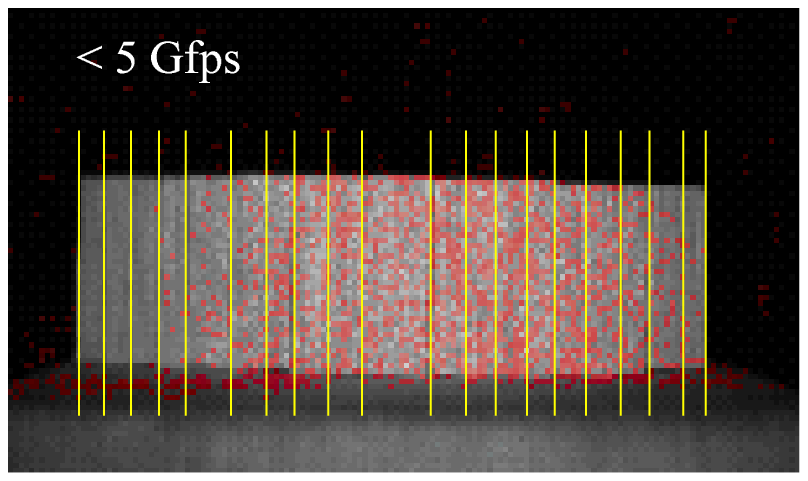}
			\label{fig:qualification_effective_fps_1x}
		}
		\subfloat[Temporal resolution (proposed)]{
			\includegraphics[width=1.5in]{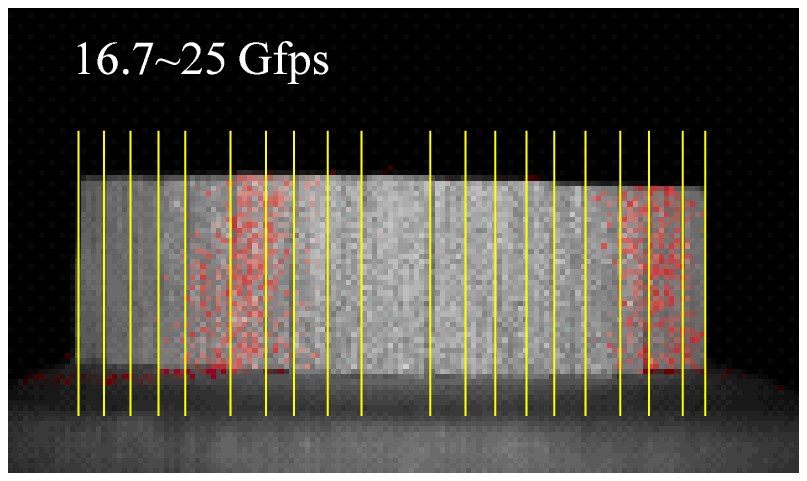}
			\label{fig:qualification_effective_fps_10x}
		}
	\end{center}
	\caption{
		(color in electronic version) {\bf Performance evaluation:} 
		(a) 10 sheets of 3 mm thickness are stacked to create the subject.
		Camera and light source are placed normal to the plane of the sheets.
		See also Fig.~\ref{fig:scene_setup}~(d).
		(b) The yellow lines indicate the boundary of each sheet. The red pixel bands occupy all the surface of the subject, which means that the effective temporal resolution of \cite{Kadambi2013} is at most 5 Gfps.
		(c) On the other hand, in our result, the red pixel bands occupy 2--3 sheets in a single frame, which correspond to 16.7--25 Gfps.
	}
	\label{fig:qualification}
\end{figure}
\Section{Results}

In this section, we show experimental results both in quantitative and qualitative manner. 
In the visualization process, we perform a simple peak detection based on Orthogonal Matching Pursuit (OMP) technique to show the wave front propagation, similar to \cite{Kadambi2013}. 
While solving OMP, we set the sparsity to one and the bases as a set of phase shifted versions of the observed kernel function.
To obtain sufficient data to perform OMP, around two thousand measurements of different $\phi$ is acquired \cite{Kadambi2013}. 
Though we only demonstrated a single path method for proof of concept, note that our temporal resolution increasing method is generalizable to multiple path methods like \cite{Heide2013,Kadambi2013,OToole2014}.

\begin{figure}[t]
	\begin{flushleft}
		\small{1x (conventional \cite{Kadambi2013}) (Yellow lines show edges of the sheets with 3 mm of thickness)}
	\end{flushleft}
	\vspace{-10pt}
	\centering\includegraphics[width=\textwidth]{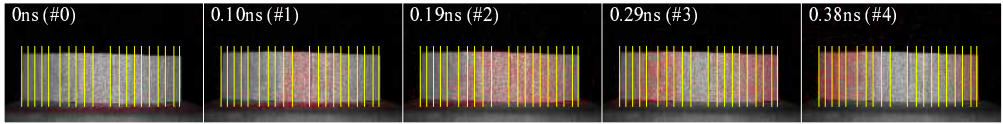}
	\vspace{-20pt}
	\begin{flushleft}
		\small{10x (with spatial phase sweep (proposed))}
	\end{flushleft}
	\vspace{-10pt}
	\centering\includegraphics[width=\textwidth]{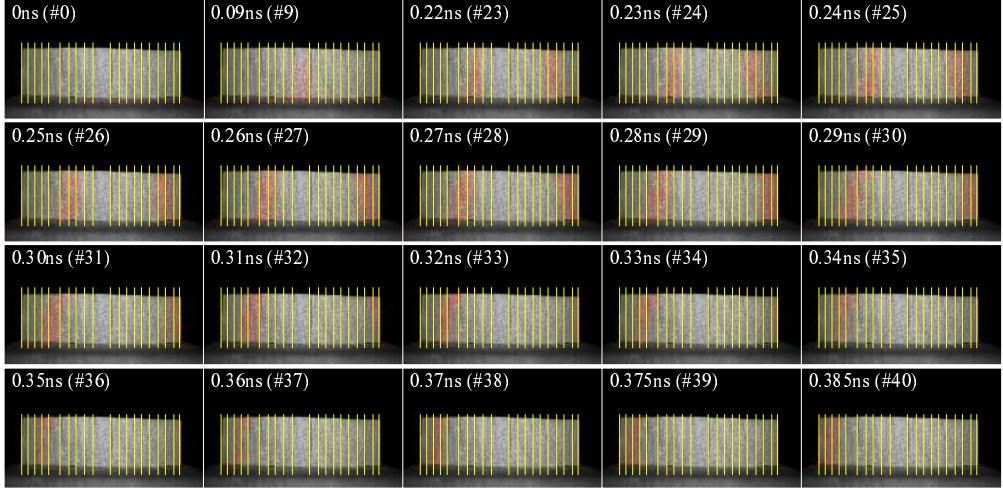}
	\caption{
		{\bf Quantitative results:} 
		The array of images shows the successive frames of transient image. 
		Images above the dotted line correspond to the result without spatial phase sweep (original frame rate determined by the PLL phase shift capability) and the ones below the line corresponds to the results of our method. 
		Red pixels indicate that the light reaches those positions at the timing of corresponding frames. 
		The elapsed time and frame index are shown in the top left of each frame. 
		Our result resolves the transient phenomenon into 40 frames which makes the width of the band of red pixels narrower than conventional transient image \cite{Kadambi2013}, which resolve the same phenomenon in only 4 steps.
		See also Visualization 1 and 2 for video version.
		}
	\label{fig:quantitative_result}
\end{figure}

\Subsection{Effective temporal resolution}
\label{sec:performance}
In this section, we experimentally evaluate the increase in temporal resolution by our method. 
We placed a terraced slope with 3 mm thick sheets arranged in front of the camera as shown in Fig.~\ref{fig:qualification}~\subref{fig:qualification_evaluation_setup}. 
We quantify the temporal resolution as the number of sheets occupied by the wavefront as shown in Fig.~\ref{fig:qualification}~\subref{fig:qualification_effective_fps_10x}.
The reconstructed wave front of light propagation of the state-of-the-art (1x) and our technique (10x) is shown in Fig.~\ref{fig:quantitative_result}. 
We can clearly observe the significantly improved temporal resolution of our spatial phase-sweep.

The effective temporal resolution can be measured from the width of the red pixel band. 
We pick up a frame and count the number of sheets the red pixel band occupies. 
Suppose the band occupies $n$ sheets, the temporal resolution of the transient image in Frames Per Second ($FPS$) will be $(3.0\times 10^{8})/(0.003\times n\times 2)$. 
The factor 2 is added as the frame rate appears doubled because of the distance traveled by light from light source to camera via the subject. 
From Fig.~\ref{fig:qualification}~\subref{fig:qualification_effective_fps_10x}, the band lies in 2--3 sheets in a single frame, which turns into 16.7--25 Gfps.
The size of the subject is too small to tell the effective temporal resolution of conventional transient image in the same manner.
However, temporal resolution is at most 5 Gfps, since the whole 10 sheets are occupied by the red pixel band in a single frame.

\begin{figure}[t]
	\begin{center}
		\subfloat[Depth maps.]{
			\includegraphics[width=2.7in]{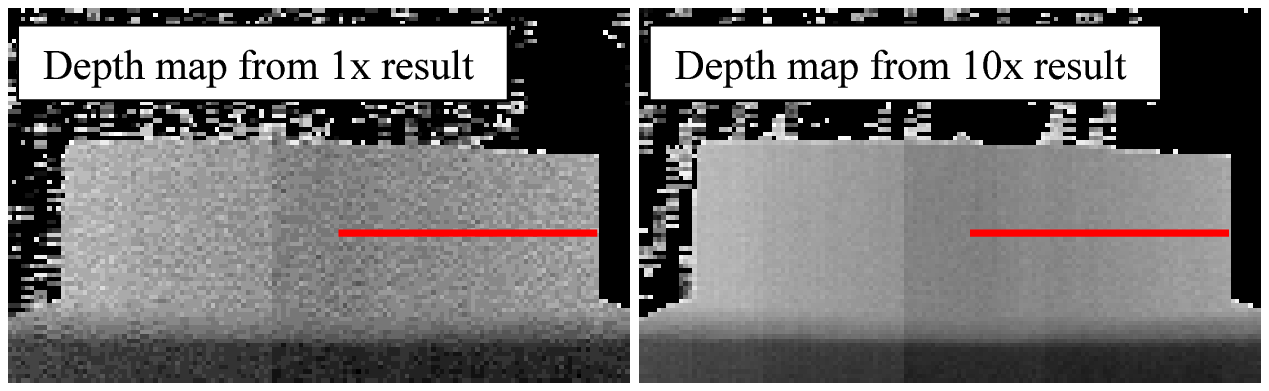}
		}
		\subfloat[Plotted depth reconstruction.]{
			\includegraphics[width=2.4in]{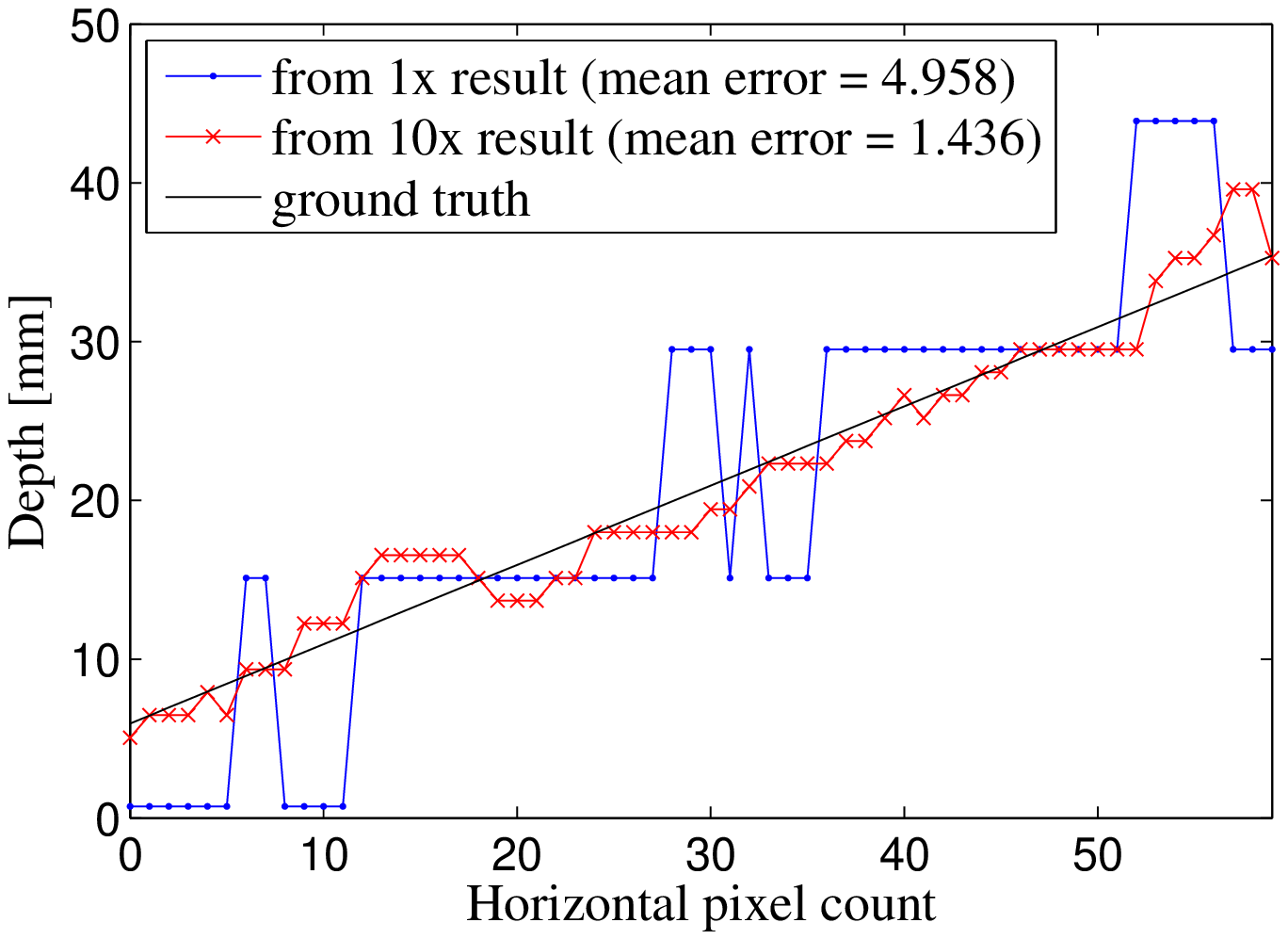}
		}
	\end{center}
	\caption{
		{\bf Depth reconstruction comparison:}
		From 1x and 10x results, we generated:
		(a) depth maps and
		(b) plot a part of it for comparison.
		The depth reconstruction graph plots along the red lines that are indicated in the depth maps.
		The mean error from the ground truth for 1x and 10x results are 4.96 mm and 1.44 mm respectively.
		To decrease the effect of noise, we applied 5 taps of median filter to the data. 
		}
	\label{fig:depth_reconstruction}
\end{figure}

Our spatial phase sweep technique also improves the accuracy of the depth estimate. 
To illustrate that, we generate depth maps using the data in Fig.~\ref{fig:quantitative_result} and plotted them as shown in Fig.~\ref{fig:depth_reconstruction}. 
We can observe that 10x result resolves the object's depth values into 20 uniform levels whereas 1x resolves to only 2 uniform levels for the same depth range.

\Subsection{Light propagation on tiny objects}
We have evaluated the effect of our technique on several scenes that includes tiny objects, small enough to describe the improvement by the proposed method.
The setups are shown in Fig.~\ref{fig:scene_setup}.

\begin{figure}[h]
	\begin{flushleft}
		\small{1x (conventional \cite{Kadambi2013})}
	\end{flushleft}
	\vspace{-10pt}
	\centering\includegraphics[width=\textwidth]{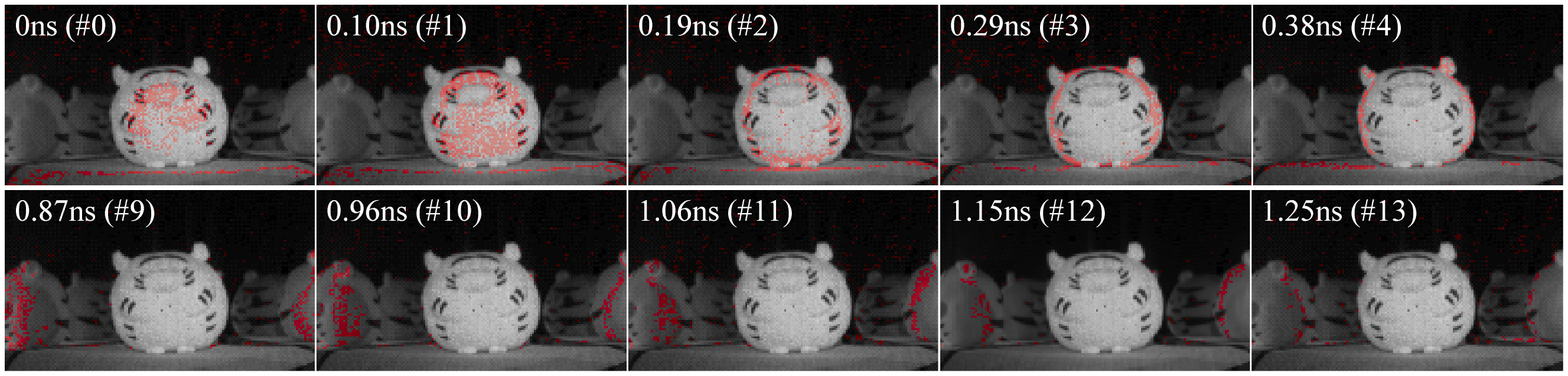}
	\vspace{-20pt}
	\begin{flushleft}
		\small{10x (with spatial phase sweep (proposed))}
	\end{flushleft}
	\vspace{-10pt}
	\centering\includegraphics[width=\textwidth]{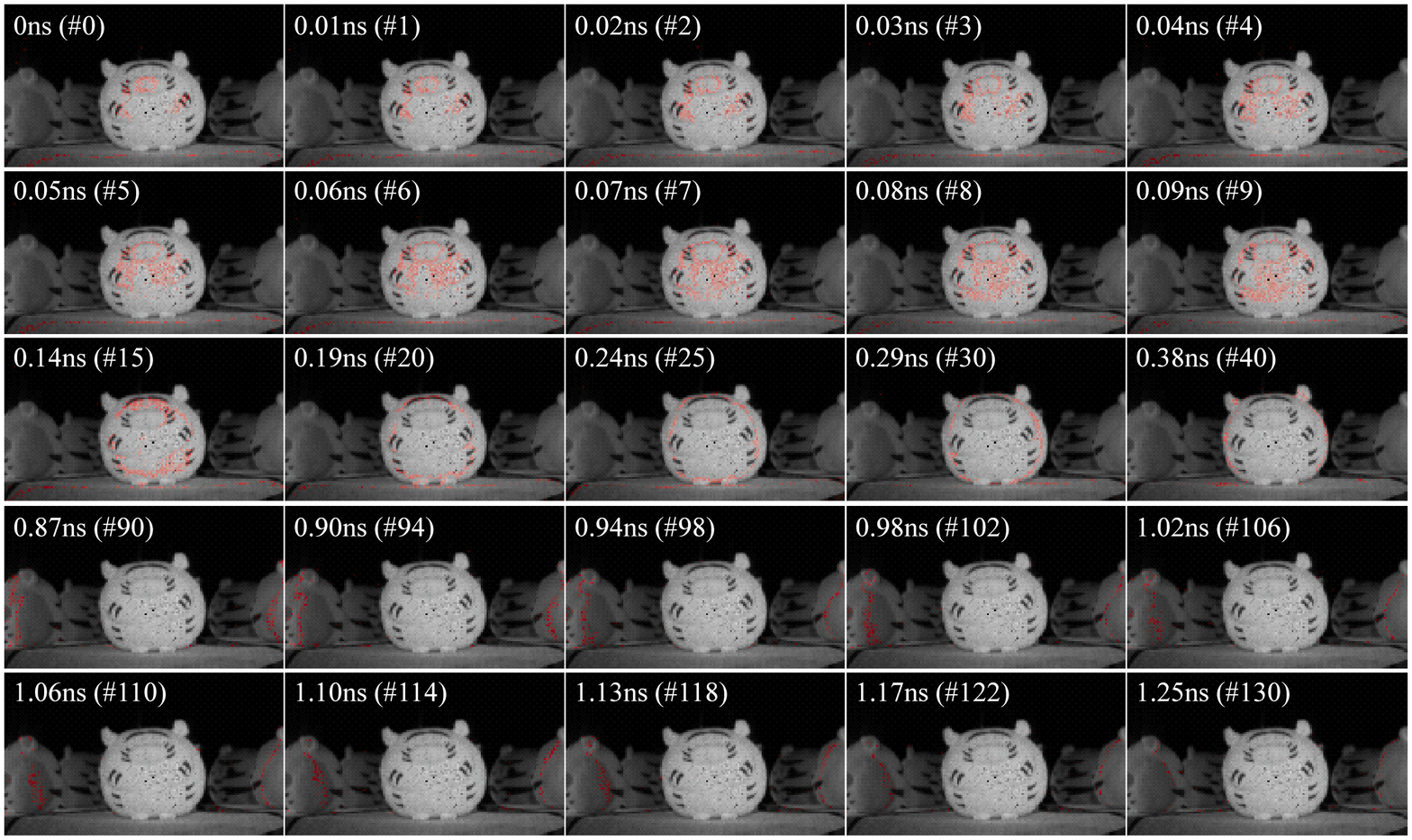}
	\caption{
		{\bf Coupled mirror scene results:} 
		Images above the dotted line correspond to the result without spatial phase sweep and the ones below the line correspond to the results of our method.
		The wave front propagation is captured more precisely in the 10x result compared to 1x. The same phenomenon, which occurs within 1--2 frames (\#0--\#2) in the 1x result, is resolved into 10--20 frames (\#0--\#20) in the 10x result.
		See also Visualization 3 and 4 for video version.
		}
	\label{fig:result_coupled_mirror}
\end{figure}

\paragraph*{Coupled mirror:}
Consider the set up in Fig.~\ref{fig:scene_setup}~(b). 
The transient images are shown in Fig.~\ref{fig:result_coupled_mirror}. The effects of 1x and 10x are similar to the quantification experiment.
Consider the top row of 1x and top three rows of 10x results. 
We can notice that the propagating wave front of the light on the stuffed toy's surface is resolved more precisely in the 10x result. 
The light hits at its nose and arms first, then gradually propagates onto its stomach and forehead taking 10--20 frames in the 10x result. 
On the other hand, the same phenomenon occurs within only 1--2 frames in the 1x result. 
The width of the band of red pixels is narrower in 10x than 1x. 
Note that the wave front moves from outside in the last half of the sequence because of the imaginary light sources created by the mirror (recall Fig.~\ref{fig:scene_setup}~(b)).

\begin{figure}[th]
	\begin{flushleft}
		\small{1x (conventional \cite{Kadambi2013})}
	\end{flushleft}
	\vspace{-10pt}
	\centering\includegraphics[width=\textwidth]{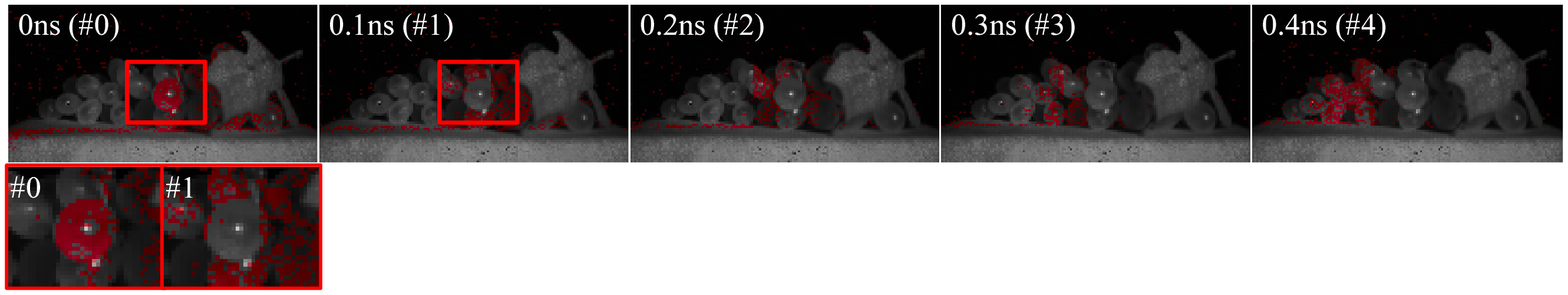}
	\vspace{-20pt}
	\begin{flushleft}
		\small{10x (with spatial phase sweep (proposed))}
	\end{flushleft}
	\vspace{-10pt}
	\centering\includegraphics[width=\textwidth]{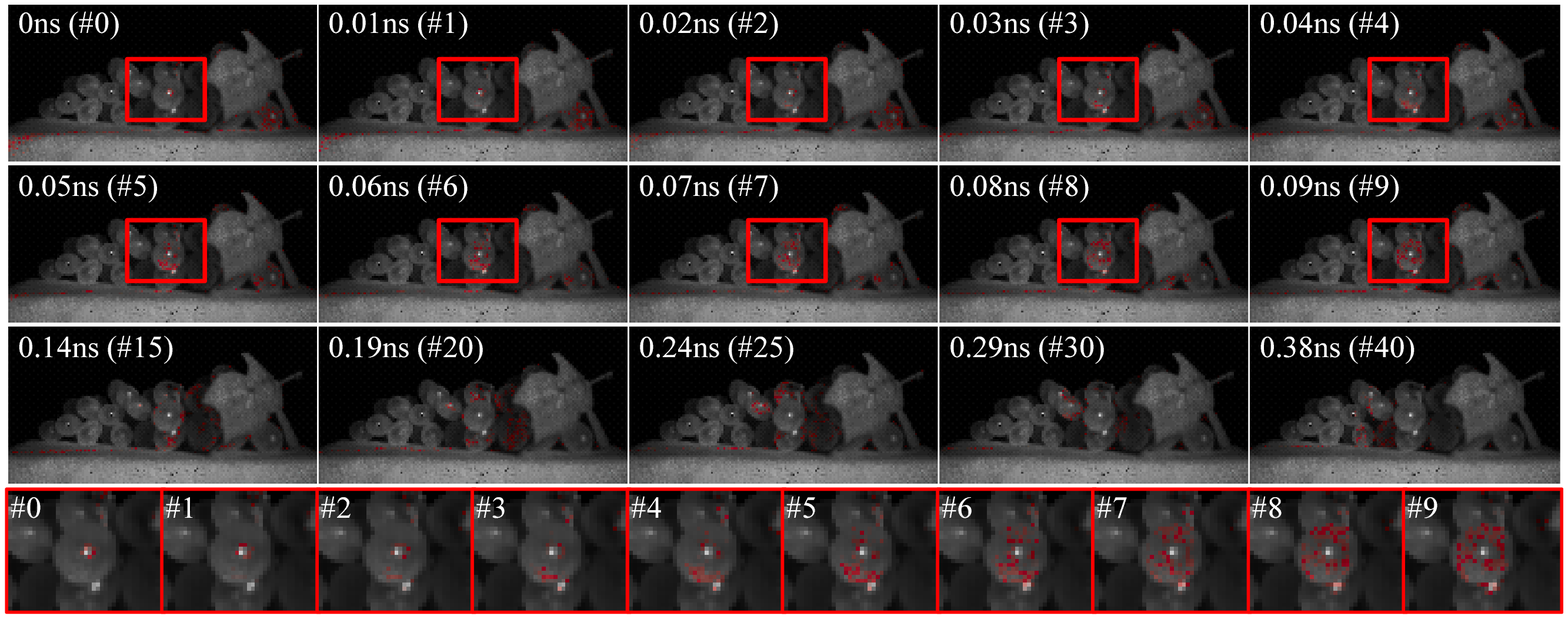}
	\caption{
		{\bf Grapes scene results:}
		Images above the dotted line correspond to the result without spatial phase sweep and the ones below the line correspond to the result of our method.
		10x result resolves the way light propagates even on a single grape while 1x result takes only 1 frame to cover each grape.
		See also Visualization 5 and 6 for video version.
		}
	\label{fig:result_grapes}
\end{figure}
\textbf{Grapes scene:}
Fig.~\ref{fig:result_grapes} shows the transient imaging result of set up in Fig.~\ref{fig:scene_setup}~(c).
The light source is placed on the right side of the scene. 
Although we can infer that light is traveling from right side to left side in both of 1x and 10x results, it can be noticed that 10x result describes the phenomenon more precisely than 1x result. 
In 10x result, we can observe the light propagation even on a single grape.


\textbf{Hue colorization:}
\begin{figure}[th]
	\centering\includegraphics[width=5in]{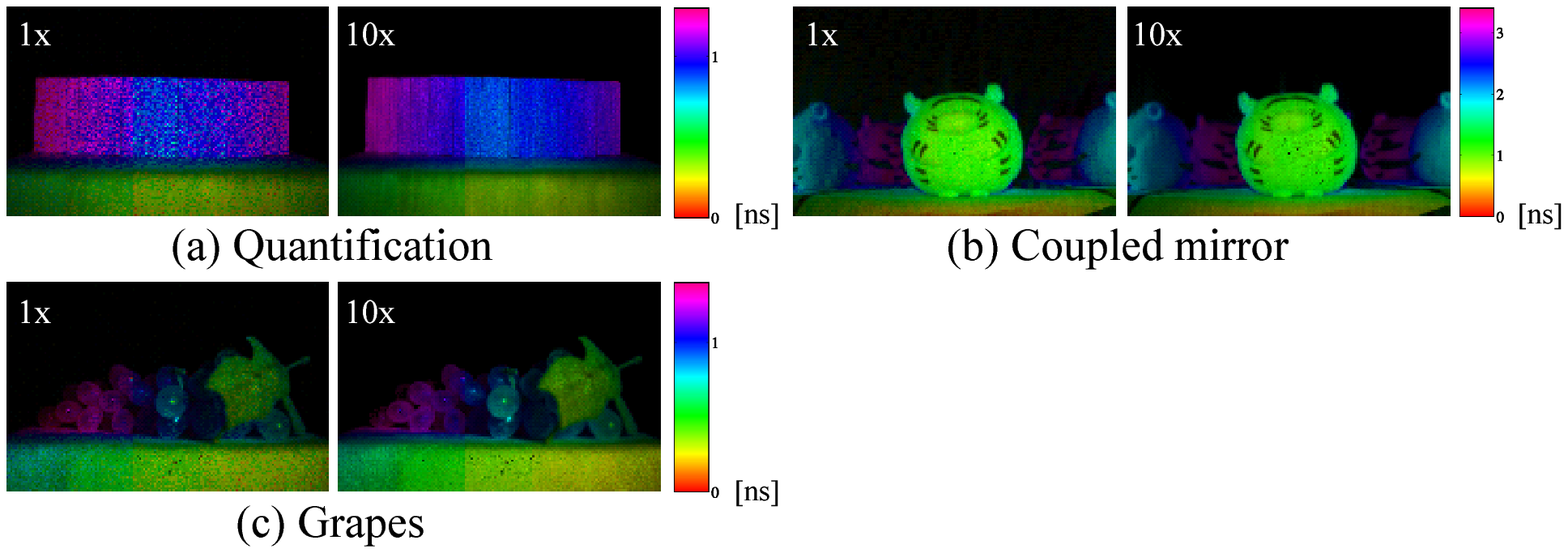}
	\caption{
		{\bf Hue colorization:}
		Hue colorized visualizations are shown using the same data as the results above.
		(a) Quantification, (b) coupled mirror, and (c) grapes.
		Left images correspond to the 1x result and right images correspond to the 10x result.
		We can notice that while the temporal resolution of 1x result is too small to represent scene response using the all color indicated in the color bar, 10x results illustrates the transient image in color smoothly.
	}
	\label{fig:hue_images}
\end{figure}
In Fig.~\ref{fig:hue_images}, we show the hue colorized visualization of the transient images using the same data.

\Section{Discussion and conclusion}
\noindent \textbf{A simple but effective modification:}
We have demonstrated that we can increase the temporal resolution of transient imaging dramatically, by a factor of 10, using just a light source array. The light source array does not increase the cost of the setup significantly.

\noindent \textbf{Frame rate of transient image:}
Although, we have empirically shown that our method improves the temporal resolution of PMD based transient imaging system by a factor of $10$, the practical temporal resolution of our system is around 16.7-25 Gfps (see Sec.~\ref{sec:performance}).
On the other hand, the actual amount of phase delay between each successive measurement, or temporal resolution, is 9.6 ps. 
If we calculate the frame rate using the definitions used in other papers \cite{Kadambi2013,OToole2014}, the frame rate translates to 104 Gfps.
One of the possible reasons of this gap between effective and theoretical temporal resolution is the SNR of the measured correlation.
In our OMP based peak detection algorithm, signal noise in the correlation could introduce variance into the detected peak positions. 
Another possible reason is subsurface scattering effect. 
Although we obtained OMP kernel from actual data to include the subsurface scattering effect in it, the shape of the kernel loses high frequencies due to subsurface scattering and that negatively affects the OMP based peak detection.

\noindent \textbf{Limitations:}
The physical size of the light sources can limit the size of the light source array and hence, the resolution of spatial phase-sweep. 
The size of the camera, may increase due to additional light sources. 
However, this is not a limiting factor for many practical applications.
Our solution is feasible today as the phase control interval in spatial domain is 3 cm.
If that value is too large, for example several tens of meters, it would have been impractical to make such a large light source array. 
On the other hand, if the advances in electronics push the phase control of PLL to around 1 ps, then it translates to building a light source array of size 300 $\mu$m, which may not be feasible.

In terms of number of measurements, the data acquisition time of our method increases linearly with the increase in temporal resolution. This can be another limiting factor in increasing the number of light sources.



\noindent \textbf{Future directions:}
\begin{tight_itemize}
	\item {\bf Designing of light source array:}
	In this paper, we used a single light source and a translation stage. By changing the position of the translation stage we simulated the effect of a light source array.
	This requires us to adjust the translation stage at every measurement and is time consuming. 
	Designing light source array will make the system more compact and will reduce the manual work. 
	\item {\bf Decreasing number of measurements:}
	Currently, in our method the number of measurements required increases linearly with the increase of sampling rate in phase domain. As some previous works have already shown, the impulse response of the scene is sparse even if it includes multiple paths or scattering. Employing compressive sensing theory might help us to reduce the number of measurements dramatically.
	\item {\bf Advanced signal model:}
	The effective temporal resolution of light propagation is limited by the peak detection method we used. Although OMP gives us good results, more advanced signal models such as exponentially modified Gaussians \cite{Heide2014} might narrow down the variance of the wave front of light.
\end{tight_itemize}


\noindent \textbf{Conclusion:}
In this paper, we proposed a technique to increase the temporal resolution of transient imaging by translating temporal domain sampling to spatial domain sweeping. 
Theoretically, we have derived the conditions required to align light source uniformly and to make calibration simple.
Though we do a simple modification to existing PMD based transient imaging system, we have demonstrated that our method improves temporal resolution dramatically in several scenes.

\Section{Acknowledgments}
Most of the hardware and software design of our system is provided by Achuta Kadambi \cite{Kadambi2013}. 
We are extremely grateful for the detailed documentation and the in-depth instructions provided by Kadambi \etal that allowed us to build our prototype.
This work was partially supported by Sony Corporation and by NSF Grants IIS:1116718 and CCF:1117939.

\begin{figure}[t]
	\begin{center}
		\subfloat[Overview]{
			\includegraphics[width=1.7in]{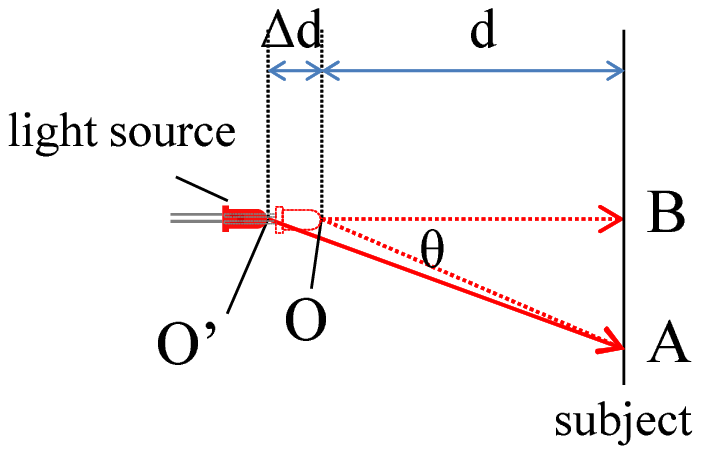}
			\label{fig:systematic_error_overview}
			}
			\hspace{0.3in}
			\subfloat[Maximum systematic error]{
				\includegraphics[width=3.0in]{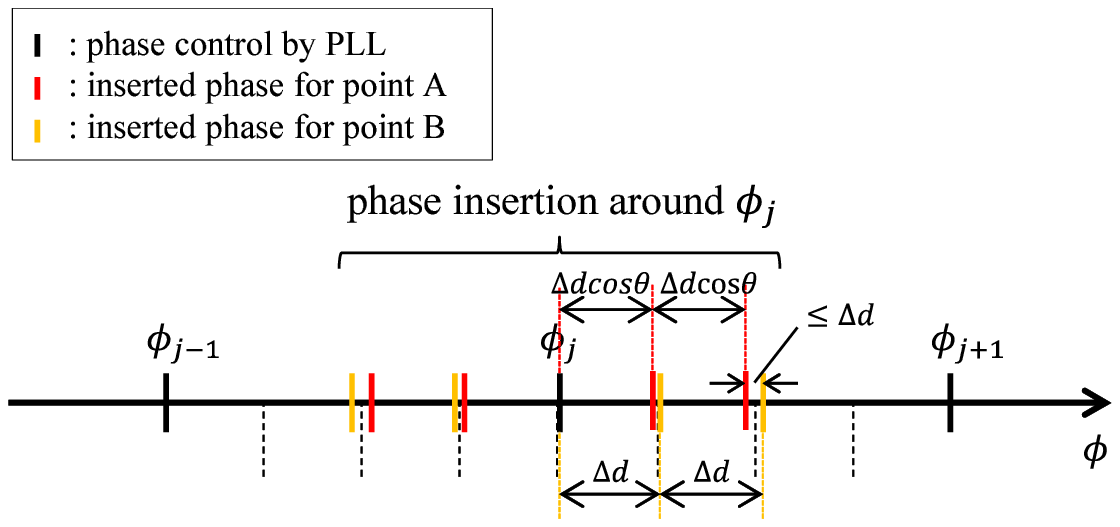}
				\label{fig:systematic_error_max}
				}
	\end{center}
	\caption{
		\textbf{Systematic error:}
		(a) $d$ is the distance between subject and light source, $\Delta d\,(\ll d)$ is the distance between two consecutive light sources and $\theta$ is $\angle AOB$. 
		Systematic error is computed in Eq.~\ref{eq:error_def}.
		(b) 
		Black solid lines indicate phases controlled by PLL, which are same for all pixels.
		Red and orange lines are phases inserted for point A and B respectively around $\phi_{j}$ by the light source array.
		The systematic error is the difference between red and orange lines.
		The maximum error happens at furthermost inserted phase from the black solid lines and is given by $\floor*{\frac{N}{2}} \Delta d (1-\cos\theta)$, where $N$ is the size of the light source array.
	}
	\label{fig:systematic_error}
\end{figure}

\appendix
\def\thesection{Appendix \Alph{section}}
\Section{Systematic error analysis of phase insertion} 
\label{subsection:Systematicerror}
In the case of single path scene, the amount of phase insertion is a function of the angle between the axis of the light source array and light ray to the subject. 
This angle dependence of spatial phase-sweep is illustrated in Fig.~\ref{fig:systematic_error}~\subref{fig:systematic_error_overview}. 
Each pixel has a different amount of phase insertion when the light source is changed. 
We define {\it systematic error} as the maximum difference in the phase shifts introduced to all the pixels by the change in the light source. 
It is possible to account for these differences in the calibration process by estimating the angle for each pixel. 
However, this demands additional calibration steps to obtain such an information. 
Hence, to keep the system simple, we will evaluate the systematic error theoretically to find the limit on the number of light sources below which, we can neglect the phase difference between pixels in the same frame. 


Consider a simple situation where we have a planar subject and the light sources are perpendicular to the subject as shown in Fig.~\ref{fig:systematic_error}.
We will first calculate the amount of phase shift introduced for a point A, $S=\mathrm{|O'A|-|OA|}$ as a function of $\theta$ and then find the systematic error by maximizing the difference between two farthest points A and B. 
For simplicity, let us assume that B is on the line of light source array. 
Hence, $\mathrm{|O'B|-|OB|}=\Delta d$. 
Using primal trigonometry, $S$ and its 1st order Maclaurin expansion can be written in terms of $\Delta d$ as follows:
\begin{equation}
S=\sqrt{\frac{d^{2}}{\cos^2\theta}+2d\Delta d+\Delta d^{2}}-\frac{d}{\cos\theta}\simeq\Delta d\cos\theta
\label{eq:error_def}
\end{equation}
with a remainder (error) term:
\begin{equation}
|R_{2}|=\left|\frac{\alpha^{2}-\frac{\beta^{2}}{4}}{2(\alpha^{2}+\beta c+c^{2})^{\frac{3}{2}}}\Delta d^{2}\right|\le\left|\frac{\alpha^{2}-\frac{\beta^{2}}{4}}{2\alpha^{3}}x^{2}\right|
\label{eq:remainder_term}
\end{equation}
where $\alpha=d/\cos\theta$, $\beta=2d$ are substitution variables and $R_{2}$ is the remainder term for 1st order Mclaurin expansion. 
According to Eq.~\ref{eq:error_def}, the difference in the inserted phases for point A and B is proportional to the amount of phase shift introduced ($\Delta d$). 
Suppose we want to increase the temporal resolution by N times. 
We make sure that the worst phase inserted does not deviate from the ideal phase by the amount of phase shift introduced. 
Hence, we have
\begin{eqnarray}
\floor*{\frac{N}{2}} \Delta d (1-\cos\theta) \le \Delta d \Rightarrow \floor*{\frac{N}{2}} \le \frac{1}{1-\cos\theta}
\end{eqnarray}
\noindent where $\Delta d$ indicates the minimum distance between the light sources below which the pixels in the same measurement can be considered to have same phase. 
This is illustrated in Fig.~\ref{fig:systematic_error}\subref{fig:systematic_error_max}. 
Suppose that the illuminated range is less than $50^\circ$ ($\theta\le25^\circ$), like a normal lens, the maximum magnification in temporal resolution will be: $N\le21.3$.
From Eq.~\ref{eq:remainder_term}, we can evaluate the accuracy of the approximation in Eq.~\ref{eq:error_def}. 
As mentioned in Sec.~\ref{sec:experiments}, the scale of our setup is as follows: $d\ge0.1\,\mathrm{m}$, $\theta\le25^\circ$ and $\Delta d\le0.03\,\mathrm{m}$. 
Then the approximation error is less than $7.3\times10^{-4}\,\mathrm{m}$. 
This result shows that the approximation in Eq.~\ref{eq:error_def} is sufficient.



\end{document}